\documentclass[11pt]{article}
\usepackage[final]{acl}

\usepackage{times}
\usepackage{latexsym}
\usepackage{booktabs}
\usepackage{tabularx}
\usepackage{subcaption}
\usepackage{comment}
\usepackage{amssymb}
\usepackage{natbib}
\usepackage[T1]{fontenc}

\usepackage[utf8]{inputenc}

\usepackage{microtype}

\usepackage{inconsolata}

\usepackage{graphicx}
\usepackage{placeins}

\usepackage{algorithm}
\usepackage{algpseudocode} 
\usepackage{amsmath} 
\usepackage{enumitem}
\usepackage{xcolor}
\usepackage{tcolorbox}

\usepackage{tabularx}
\usepackage{colortbl}

\usepackage{booktabs}

\tcbuselibrary{breakable}

\newcommand{\Func}[1]{\textsc{#1}}
\usepackage{multirow}

\title{MAGMA: A Multi-Graph based Agentic Memory Architecture for AI Agents}

\author{
Dongming Jiang$^{\alpha}$, Yi Li$^{\alpha}$, Guanpeng Li$^{\beta}$, Bingzhe Li$^{\alpha}$\thanks{Corresponding author} \\
$^{\alpha}$Department of Computer Science, The University of Texas at Dallas \\
$^{\beta}$Department of Electrical and Computer Engineering, University of Florida \\
\texttt{\{dongming.jiang, yi.li3, bingzhe.li\}@utdallas.edu} \\\texttt{liguanpeng@ufl.edu}
}

\begin{document}
\maketitle

\begin{abstract}

Memory-Augmented Generation (MAG) extends Large Language Models with external memory to support long-context reasoning, but existing approaches largely rely on semantic similarity over monolithic memory stores, entangling temporal, causal, and entity information. This design limits interpretability and alignment between query intent and retrieved evidence, leading to suboptimal reasoning accuracy.
In this paper, we propose MAGMA, a multi-graph agentic memory architecture that represents each memory item across orthogonal semantic, temporal, causal, and entity graphs. MAGMA formulates retrieval as policy-guided traversal over these relational views, enabling query-adaptive selection and structured context construction. By decoupling memory representation from retrieval logic, MAGMA provides transparent reasoning paths and fine-grained control over retrieval. Experiments on LoCoMo and LongMemEval demonstrate that MAGMA consistently outperforms state-of-the-art agentic memory systems in long-horizon reasoning tasks. The open-source code is available at: \url{https://github.com/FredJiang0324/MAGMA}.

\end{abstract}
\section{Introduction}
\label{sec:intro}
Large Language Models (LLMs) have demonstrated remarkable capabilities across a wide range of tasks~\cite{Brown2020,achiam2023gpt,Wei2022}, 
yet they remain limited in their ability to maintain and reason over long-term context. 
These models process information within a finite attention window, and their internal representations do not persist across interactions, 
causing earlier details to be forgotten once they fall outside the active context~\cite{Brown2020,Beltagy2020}. 
Even within a single long sequence, attention effectiveness degrades with distance due to \textit{attention dilution}, positional encoding limitations, and token interference, leading to the well-known ``lost-in-the-middle'' and context-decay phenomena~\cite{liu2024lost,Press2021}. 
Moreover, LLMs lack native mechanisms for stable and structured memory, resulting in inconsistent recall, degraded long-horizon reasoning, and limited support for tasks requiring persistent and organized memory~\cite{Khandelwal2018,maharana2024evaluating}.

To address these inherent limitations, Memory-Augmented Generation (MAG) systems have emerged as a promising direction for enabling LLMs to operate beyond the boundaries of their fixed context windows. MAG equips an agent with an external memory continuously recording interaction histories and allowing the agents to retrieve and reintegrate past experiences when generating new responses. By offloading long-term context to an explicit memory module, MAG systems provide a means for agents to accumulate knowledge over time, support multi-session coherence, and adapt to evolving conversational or task contexts. In this paradigm, memory is no longer implicit in internal activations but becomes a persistent, queryable resource that substantially enhances long-horizon reasoning, personalized behavior, and stable agent identity.

Despite their promise, current MAG systems exhibit structural and operational limitations that constrain their effectiveness in long-term reasoning~\cite{li2025memos, chhikara2025mem0, xu2025mem, packer2023memgpt, rasmussen2025zep, wang2025mirix, Kang2025}. Most existing approaches store past interactions in monolithic repositories or minimally structured memory buffers, relying primarily on semantic similarity, recency, or heuristic scoring to retrieve relevant content. For example, A-Mem~\cite{xu2025mem} organizes past interactions into Zettelkasten-like memory units that are incrementally linked and refined, yet their retrieval pipelines rely primarily on semantic embedding similarity, missing the relations such as temporal or causal relationships. Cognitive-inspired frameworks like Nemori~\cite{nan2025nemori} introduce principled episodic segmentation and representation alignment, enabling agents to detect event boundaries and construct higher-level semantic summaries. However, their memory structures are still narrative and undifferentiated, with no explicit modeling of distinct relational dimensions.

To address the structural limitations of existing MAG systems, we propose MAGMA, a multi-graph agentic memory architecture that explicitly models heterogeneous relational structure in an agent’s experience. MAGMA represents each memory item across four orthogonal relational graphs (i.e., semantic, temporal, causal, and entity), yielding a disentangled representation of how events, concepts, and participants are related.

Built on this unified multi-graph substrate, MAGMA introduces a hierarchical, intent-aware query mechanism that selects relevant relational views, traverses them independently, and fuses the resulting subgraphs into a compact, type-aligned context for generation. By decoupling memory representation from retrieval logic, MAGMA enables transparent reasoning paths, fine-grained control over memory selection, and improved alignment between query intent and retrieved evidence. This relational formulation provides a principled and extensible foundation for agentic memory, improving both long-term coherence and interpretability.

Our contributions are summarized as follows:
\begin{enumerate}[leftmargin=*, itemsep=0pt, topsep=2pt]
    \item We propose \textbf{MAGMA}, a multi-graph agentic memory architecture that explicitly models semantic, temporal, causal, and entity relations essential for long-horizon reasoning.
    \item We introduce an Adaptive Traversal Policy that routes retrieval based on query intent, enabling efficient pruning of irrelevant graph regions and achieving lower latency and reduced token usage.
    \item We design a dual-stream memory evolution mechanism that decouples latency-sensitive event ingestion from asynchronous structural consolidation, preserving responsiveness while refining relational structure.
    \item We demonstrate that MAGMA consistently outperforms state-of-the-art agentic memory systems on long-context benchmarks including LoCoMo and LongMemEval, while reducing retrieval latency and token consumption relative to prior systems. The code is open-sourced\footnote{https://github.com/FredJiang0324/MAGMA}.
\end{enumerate}

\begin{figure}[t]
    \centering
    \includegraphics[width=0.85\linewidth]{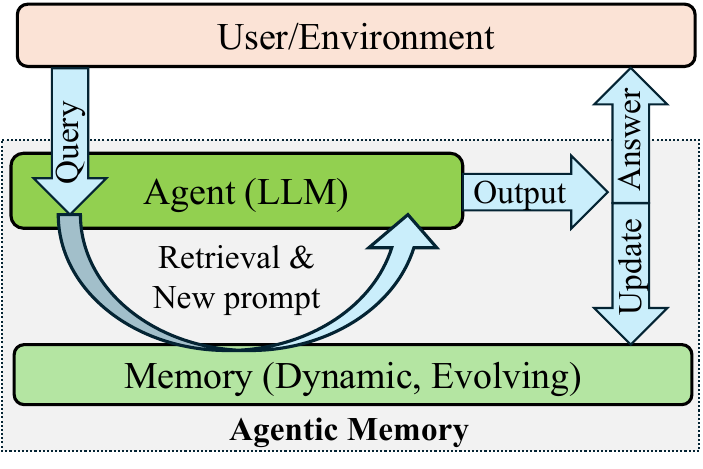}
    \caption{High-Level Architecture of Memory-Augmented Generation (MAG).}
    \label{fig:mag_overview}
\end{figure}

\begin{figure*}[t]
    \centering
    \includegraphics[width=0.8\linewidth]{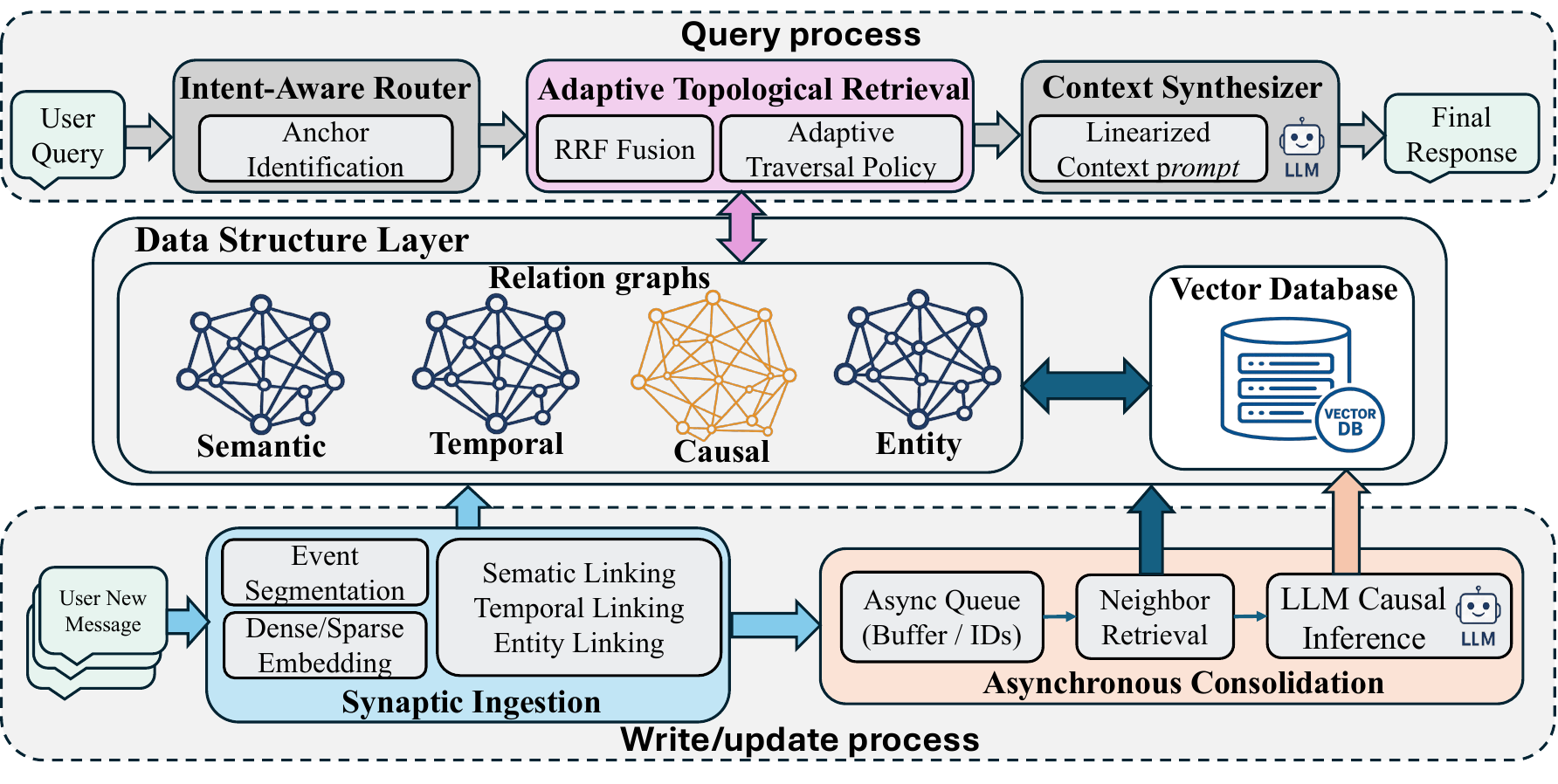}
    \caption{Architectural Overview of MAGMA. The system is composed of three layers: (1) A Query Process that routes and synthesizes context; (2) A Data Structure Layer organizing memory into Relation Graphs and a Vector Database; and (3) A Write/Update Process utilizing a dual-stream mechanism for fast ingestion and asynchronous consolidation.}
    \label{fig:MAGMA_overview}
\end{figure*}

\section{Background}
\label{sec:related}
Existing Large Language Models (LLMs) face fundamental challenges in handling long-term agentic interactions. These challenges stem from the inherent limitations of fixed-length contexts, which result in fragmented memory and an inability to maintain narrative coherence over time. 
The evolution of long-term consistency in LLMs is shifted from \textit{Context-Window Extension}~\citep{Beltagy2020,Press2021,kang2025lm2,qian2025memorag}, \textit{Retrieval-Augmented Generation (RAG)}~\citep{lewis2020retrieval,jiang2025rago,wang2024m,jiang2024longrag,gutierrez2025rag,lin2025cache} to \textit{Memory-Augmented Generation (MAG)}.

Retrieval-oriented approaches enrich the model with an external, dynamic memory library, giving rise to the paradigm of Memory-Augmented Generation (MAG)~\cite{zhong2024memorybank, park2023generative, huang2024emotional}. Formally, unlike static RAG, MAG maintains a time-variant memory $\mathcal{M}_t$ that evolves via a feedback loop:
\begin{equation}
    o_t = \text{LLM}(q_t, \text{Retrieve}(q_t, \mathcal{M}_t))
\end{equation}
\begin{equation}
    \mathcal{M}_{t+1} = \text{Update}(\mathcal{M}_t, q_t, o_t)
\end{equation}

As shown in Figure~\ref{fig:mag_overview}, this feedback loop enables the memory module to evolve over time: the user query is combined with retrieved information to form an augmented prompt, and the model’s output is subsequently written back to refine $\mathcal{M}_t$.


Some prior schemes focused on structuring the intermediate states or relationships of memory to enable better reasoning. 
Think-in-Memory (TiM) \citep{liu2023think} stores evolving chains-of-thought to maintain consistency. 
A-MEM \citep{xu2025mem} draws inspiration from the Zettelkasten method, organizing knowledge into an interconnected note network. 
More recently, graph-based approaches like GraphRAG \citep{edge2024graphrag} and Zep \citep{rasmussen2025zep} structure memory into knowledge graphs to capture cross-document dependencies. We provide a detailed discussion of related work in Appendix~\ref{app:relatedwork}.

However, prior work typically organizes memory around associative proximity (e.g., semantic similarity) rather than mechanistic dependency~\citep{kiciman2023causal}. As a result, such methods can retrieve \textit{what} occurred but struggle to reason about \textit{why}, since they lack explicit representations of causal structure, leading to reduced accuracy in complex reasoning tasks~\citep{jin2023cladder, zhang2025igniting}.

\section{MAGMA Design}
\label{sec:system}
In this section, we introduce the proposed Multi-Graph based Agentic Memory (MAGMA) design and its components in detail.

\subsection{Architectural Overview}
\label{sec:system:overview}

MAGMA architecture is organized into the following three logical layers, orchestrating the interaction between control logic and the memory substrate as illustrated in Figure~\ref{fig:MAGMA_overview}.

\noindent \textbf{Query Process:} The inference engine responsible for retrieving and synthesizing information. It comprises the \textit{Intent-Aware Router} for dispatching tasks, the \textit{Adaptive Topological Retrieval} module for executing graph traversals, and the \textit{Context Synthesizer} for generating the final narrative response.
    
\noindent \textbf{Data Structure ($\mathcal{G}$):} The unified storage substrate that fuses disparate modalities. As shown in the center of Figure~\ref{fig:MAGMA_overview}, it maintains a \textit{Vector Database} for semantic search alongside four distinct \textit{Relation Graphs} (i.e., Semantic, Temporal, Causal and Entity). This layer provides the topological foundation for cross-view reasoning.

\noindent  \textbf{Write/Update Process:} A dual-stream pipeline manages memory evolution. It decouples latency-sensitive operations via \textit{Synaptic Ingestion} (Fast Path) from compute-intensive reasoning via \textit{Asynchronous Consolidation} (Slow Path), ensuring the system remains responsive while continuously deepening its memory structure.

Functionally, the Query Layer interacts with the Data Structure Layer to execute the synchronous Query Process (Section~\ref{sec:system:query}), while the Write/Update Layer manages the continuous Memory Evolution (Section~\ref{sec:system:update}).

\subsection{Data Structure Layer}
\label{sec:system:topology}
As the core component of Memory-Augmented Generation (MAG), the data structure layer is responsible for storing, organizing, and evolving past information to support future retrieval and updates. In MAGMA, we formalize this layer as a time-variant directed multigraph $\mathcal{G}_t = (\mathcal{N}_t, \mathcal{E}_t)$, where nodes represent events and edges encode heterogeneous relational structures. This unified manifold enables structured reasoning across multiple logical dimensions(i.e., semantic, temporal, causal, and entity) while preserving their orthogonality.


\noindent \textbf{Unified node representation:}
The node set $\mathcal{N}$ is hierarchically organized to represent experience at multiple granularities, ranging from fine-grained atomic events to higher-level episodic groupings. Each Event-Node $n_i \in \mathcal{N}_{\text{event}}$ is defined as:
\begin{equation}
    n_i = \langle c_i, \tau_i, \mathbf{v}_i, \mathcal{A}_i \rangle
\end{equation}
where $c_i$ denotes the event content (e.g., observations, actions, or state changes), $\tau_i$ is a discrete timestamp anchoring the event in time, and $\mathbf{v}_i \in \mathbb{R}^d$ is a dense representation indexed in the vector database~\citep{johnson2019billion}. The attribute set $\mathcal{A}_i$ captures structured metadata such as entity references, temporal cues, or contextual descriptors, enabling hybrid retrieval that integrates semantic similarity with symbolic and structural constraints.

\noindent \textbf{Relation graphs (edge space):}
The edge set $\mathcal{E}$ is partitioned into four semantic subspaces, corresponding to the relation graphs:
\begin{itemize}[leftmargin=*, itemsep=0pt, topsep=2pt]
    \item \textbf{Temporal Graph ($\mathcal{E}_{temp}$):} Defined as strictly ordered pairs $(n_i, n_j)$ where $\tau_i < \tau_j$. This immutable chain provides the ground truth for chronological reasoning.
    \item \textbf{Causal Graph ($\mathcal{E}_{causal}$):} Directed edges representing logical entailment. An edge $e_{ij} \in \mathcal{E}_{causal}$ exists if $S(n_j|n_i, q) > \delta$, explicitly inferred by the consolidation module to support "Why" queries.
    \item \textbf{Semantic Graph ($\mathcal{E}_{sem}$):} Undirected edges connecting conceptually similar events, formally defined by $\cos(\mathbf{v}_i, \mathbf{v}_j) > \theta_{sim}$.
    \item \textbf{Entity Graph ($\mathcal{E}_{ent}$):} Edges connecting events to abstract entity nodes, solving the object permanence problem across disjoint timeline segments.
\end{itemize}

\begin{figure*}[t]
\centering
\includegraphics[width=\textwidth]{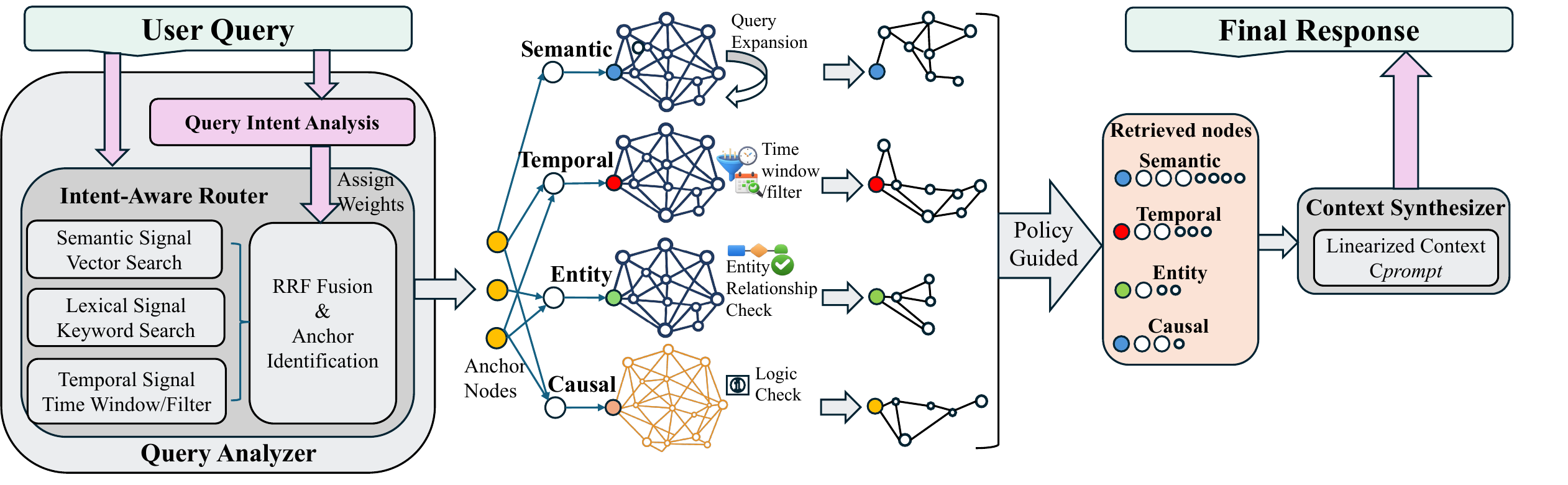}
\caption{Query process with adaptive hybrid retrieval pipeline. (1) Query Analysis detects intent and fuses signals to find Anchors. (2) {Adaptive Traversal} navigates specific graph views (Causal, Temporal) based on the policy weights.}
\label{fig:retrieval_pipeline}
\end{figure*}

\subsection{Query Process: Adaptive Hierarchical Retrieval}\label{sec:system:query}
As illustrated in Figure~\ref{fig:retrieval_pipeline}, retrieval in MAGMA is formulated as a policy-guided graph traversal rather than a static lookup operation. The query process is orchestrated by a Router $\mathcal{R}$, which decomposes the user query into structured control signals and executes a multi-stage retrieval pipeline (Algorithm~\ref{alg:query}) that dynamically selects, traverses, and fuses relevant relational views. Four main stages in the query process is introduced below:


\noindent\textbf{Stage 1 - Query Analysis \& Decomposition:}
The process begins by decomposing the raw user query $q$ into structured control signals, including semantic, lexical, and temporal cues. MAGMA then extracts three complementary representations to guide the retrieval process:
\begin{itemize}[leftmargin=*, itemsep=0pt, topsep=2pt]
    \item \textbf{Intent Classification ($T_q$):} A lightweight classifier maps $q$ to a specific intent type $T_q \in \{ \textsc{Why}, \textsc{When}, \textsc{Entity} \}$. This acts as the "steering wheel," determining which graph edges will later be prioritized (e.g., "Why" queries trigger a bias for Causal edges).
    \item \textbf{Temporal Parsing ($[\tau_s, \tau_e]$):} A temporal tagger resolves relative expressions (e.g., "last Friday") into absolute timestamps, defining a hard time window for filtering.
    \item \textbf{Representation Extraction:} The system simultaneously generates a dense embedding $\vec{q}$ for semantic search and extracts sparse keywords $q_{key}$ for exact lexical matching.
\end{itemize}

\noindent\textbf{Stage 2 - Multi-Signal Anchor Identification:}
Before initiating graph traversal, the system first identifies a set of anchor nodes that serve as entry points into the memory graph. To ensure robustness across query modalities, we fuse signals from dense semantic retrieval, lexical keyword matching, and temporal filtering using Reciprocal Rank Fusion (RRF)~\cite{Cormack2009}:

\begin{equation}
    \label{eq:rrf}
    S_{anchor} = \text{Top}_K \left( \sum_{m \in \{vec, key, time\}} \frac{1}{k + r_m(n)} \right)
\end{equation}
This ensures robust starting points regardless of query modality.

\noindent\textbf{Stage 3 - Adaptive Traversal Policy:}
Starting from the anchor set $\mathcal{S}_{anchor}$, the system expands the context using a Heuristic Beam Search. Unlike rigid rule-based traversals, MAGMA calculates a dynamic transition score $S(n_j | n_i, q)$ for moving from node $n_i$ to neighbor $n_j$ via edge $e_{ij}$. This score fuses structural alignment with semantic relevance:

\begin{equation}
\label{eq:beam_score}
\begin{split}
    S(n_j|n_i, q) = \exp \bigg( & \lambda_1 \cdot \underbrace{\phi(type(e_{ij}), T_q)}_{\substack{\text{Structural} \\ \text{Alignment}}} \\
    & + \lambda_2 \cdot \underbrace{\text{sim}(\vec{n}_j, \vec{q})}_{\substack{\text{Semantic} \\ \text{Affinity}}} \bigg)
\end{split}
\end{equation}

Here, $\text{sim}(\cdot)$ denotes the cosine similarity between the neighbor's embedding and the query embedding. The structural alignment function $\phi$ dynamically rewards edge types based on the detected query intent $T_q$:

\begin{equation}
    \phi(r, T_q) = \mathbf{w}_{T_q}^\top \cdot \mathbf{1}_r
\end{equation}

where $\mathbf{w}_{T_q}$ is an adaptive weight vector specific to intent $T_q$ (e.g., assigning high weights to \texttt{CAUSAL} edges for "Why" queries), and $\mathbf{1}_r$ is the one-hot encoding of the edge relation. 

At each step, the algorithm retains the top-$k$ nodes with the highest cumulative scores. This ensures the traversal is guided by a dual signal: strictly following the logical structure (via $\phi$) while maintaining contextual focus (via $\text{sim}$).

\begin{algorithm}[t]
\caption{Adaptive Hybrid Retrieval (Heuristic Beam Search)}
\label{alg:query}
\begin{algorithmic}[1]
\small
\Require Query $q$, Graph $G$, VectorDB $V$, Intent $T_q$
\Ensure Narrative Context $C_{out}$

\State \textcolor{gray!80}{\textit{// Phase 1: Initialization}}
\State $S_{anchor} \leftarrow \Func{RRF}(V.\Func{Search}(\vec{q}), K.\Func{Search}(q_{key}))$ \Comment{Hybrid Retrieval}
\State $CurrentFrontier, Visited \leftarrow S_{anchor}$ 
\State $\mathbf{w}_{T_q} \leftarrow \Func{GetAttentionWeights}(T_q)$

\For{$d \leftarrow 1$ \textbf{to} $MaxDepth$}
    \State $Candidates \leftarrow \Func{PriorityQueue}()$
    
    \For{$u \in CurrentFrontier$}
        \For{$v \in G.\Func{Neighbors}(u)$}
            \If{$v \notin Visited$}
                \State \textcolor{gray!80}{\textit{// Calculate transition score via Eq.~\ref{eq:beam_score}}}
                \State $s_{uv} \leftarrow \exp(\lambda_1 (\mathbf{w}_{T_q}^\top \cdot \mathbf{1}_{e_{uv}}) + \lambda_2 \text{sim}(\vec{v}, \vec{q}))$
                \State $score_v \leftarrow \text{score}_u \cdot \gamma + s_{uv}$ \Comment{Apply Decay $\gamma$}
                \State $Candidates.\Func{Push}(v, score_v)$
            \EndIf
        \EndFor
    \EndFor
    
    \State $CurrentFrontier \leftarrow Candidates.\Func{TopK}(BeamWidth)$
    \State $Visited.\Func{AddAll}(CurrentFrontier)$
    \If{$Visited.\Func{Size}() \ge Budget$} \textbf{break} \EndIf
\EndFor

\State $C_{sorted} \leftarrow \Func{TopologicalSort}(Visited, T_q)$
\State \Return $\Func{Serialize}(C_{sorted})$

\end{algorithmic}
\end{algorithm}

\noindent \textbf{Stage 4: Narrative Synthesis via Graph Linearization:}
The final phase transforms the retrieved subgraph $\mathcal{G}_{sub}$ into a coherent narrative context. MAGMA employs a structure-aware linearization protocol that preserves the relational dependencies encoded in the graph with the following three phases.

\textbf{1. Topological Ordering:} 
Raw nodes are reorganized to reflect the logic of the query. For temporal queries ($T_q=\textsc{When}$), nodes are sorted by timestamp $\tau_i$. For causal queries ($T_q=\textsc{Why}$), we apply a topological sort on the causal edges $\mathcal{E}_{causal}$ to ensure causes precede effects in the prompt context.

\textbf{2. Context Scaffolding with Provenance:} 
To mitigate hallucination, each node is serialized into a structured block containing its timestamp, content, and explicit reference ID. We define the linearized context $C_{prompt}$ as:
\begin{equation}
\resizebox{0.95\linewidth}{!}{$
    C_{prompt} = \bigoplus_{n_i \in \text{Sort}(\mathcal{G}_{sub})} \left[ \texttt{<t:}\tau_i\texttt{> } n_i.content \texttt{ <ref:} n_i.id \texttt{>} \right]
$}
\end{equation}
where $\bigoplus$ denotes string concatenation.

\textbf{3. Salience-Based Token Budgeting:} 
Given a fixed LLM context window, we cannot include all retrieved nodes. We utilize the relevance scores $S(n_j|n_i, q)$ computed in Eq.~(5) to enforce a dynamic budget. Low-probability nodes are summarized into brevity codes (e.g., "...3 intermediate events..."), while high-salience nodes retain full semantic detail. 

This structured scaffold forces the LLM to act as an interpreter of evidence rather than a creative writer, significantly reducing grounding errors.

\subsection{Memory Evolution (Write and Update)}
\label{sec:system:update}
Long-term reasoning requires not only effective retrieval, but also a memory substrate that can adapt and reorganize as experience accumulates. MAGMA addresses this requirement through a structured memory evolution scheme that incrementally refines its multi-relational graph over time. Specifically, the transition from $\mathcal{G}_t$ to $\mathcal{G}_{t+1}$ is governed by a dual-stream process that decouples latency-sensitive ingestion from compute-intensive consolidation~\cite{Kumaran2016}, balancing short-term responsiveness with long-term reasoning fidelity.


\noindent\textbf{Fast path ( synaptic ingestion):}
The Fast Path operates on the critical path of interaction, constrained by strict latency requirements. It performs non-blocking operations: event segmentation, vector indexing, and updating the immutable temporal backbone ($n_{t-1} \rightarrow n_t$). As detailed in Algorithm~\ref{alg:ingest}, no blocking LLM reasoning occurs here, ensuring the agent remains responsive regardless of memory size.

\begin{algorithm}[!t]
\caption{Fast Path: Synaptic Ingestion}
\label{alg:ingest}
\begin{algorithmic}[1]
\small
\Require User Interaction $I$, Current Graph $\mathcal{G}_t$
\Ensure Updated Graph $\mathcal{G}_{t+1}$
\State $n_t \leftarrow \Func{SegmentEvent}(I)$
\State $n_{prev} \leftarrow \Func{GetLastNode}(\mathcal{G}_t)$
\State \Comment{Update Temporal Backbone}
\State $\mathcal{G}.\Func{AddEdge}(n_{prev}, n_t, \text{type}=\textsc{Temp})$
\State \Comment{Indexing}
\State $\mathbf{v}_t \leftarrow \Func{Encoder}(n_t.c)$
\State $VDB.\Func{Add}(\mathbf{v}_t, n_t.id)$
\State $Queue.\Func{Enqueue}(n_t.id)$ \Comment{Trigger Slow Path}
\State \Return $n_t$
\end{algorithmic}
\end{algorithm}

\noindent\textbf{Slow path (structural consolidation):}
Asynchronously, the slow path performs Memory Consolidation (Algorithm~\ref{alg:async}). It functions as a background worker that dequeues events and densifies the graph structure. By analyzing the local neighborhood $\mathcal{N}(n_t)$ of recent events, the system employs an LLM $\Phi$ to infer latent connections:
\begin{equation}
    \mathcal{E}_{new} = \Phi_{reason}(\mathcal{N}(n_t), \mathcal{H}_{history})
\end{equation}
This process constructs high-value $\mathcal{E}_{causal}$ and $\mathcal{E}_{ent}$ links, effectively trading off compute time for relational depth.




\begin{algorithm}[!t]
\caption{Slow Path: Structural Consolidation}
\label{alg:async}
\begin{algorithmic}[1]
\small
\State \textbf{Worker Process:}
\Loop
    \State $id \leftarrow Queue.\Func{Dequeue}()$
    \If{$id$ is \textbf{null}} \textbf{continue} \EndIf
    \State $n_t \leftarrow \mathcal{G}.\Func{GetNode}(id)$
    \State $\mathcal{N}_{local} \leftarrow \mathcal{G}.\Func{GetNeighborhood}(n_t, \text{hops}=2)$
    \State \Comment{Infer latent Causal and Entity structures}
    \State $Prompt \leftarrow \Func{Format}(\mathcal{N}_{local})$
    \State $\mathcal{E}_{new} \leftarrow \Phi_{LLM}(Prompt)$
    \State $\mathcal{G}.\Func{AddEdges}(\mathcal{E}_{new})$
\EndLoop
\end{algorithmic}
\end{algorithm}

\subsection{Implementation }
\label{sec:MAGMA:impl}
We implement MAGMA as a modular three-layer architecture designed for extensibility, scalability, and deployment flexibility. The storage layer abstracts over heterogeneous physical backends, providing unified interfaces for managing the typed memory graph, dense vector indices, and sparse keyword indices. This abstraction cleanly separates the logical memory model from its physical realization, enabling seamless substitution of storage backends (e.g., in-memory data structures versus production-grade graph or vector databases) with minimal engineering effort.

The retrieval layer coordinates the core algorithmic components, including memory construction, multi-stage ranking, and policy-guided graph traversal. It is supported by specialized utility modules for episodic segmentation and temporal normalization, which provide structured signals to downstream retrieval and traversal policies. The application layer manages the interaction loop, evaluation harnesses, and prompt construction, serving as the interface between the agent and the underlying memory system.

\begin{table*}[t]
\centering
\small
\caption{Performance on the LoCoMo benchmark evaluated using the LLM-as-a-Judge metric. Higher scores indicate better performance. LLM model is based on gpt-4o-mini.}
\label{tab:locomo-main}

\resizebox{0.8\textwidth}{!}{
\begin{tabular}{lccccc|c}
\toprule
Method &
Multi-Hop &
Temporal &
Open-Domain &
Single-Hop &
Adversarial &
Overall \\
\midrule
\midrule
Full Context &
0.468 &
0.562 &
0.486 &
0.630 &
0.205 &
0.481 \\

A-MEM &
0.495 &
0.474 &
0.385 &
0.653 &
0.616 &
0.580 \\

MemoryOS &
0.552 &
0.422 &
0.504 &
0.674 &
0.428 &
0.553 \\

Nemori &
\textbf{0.569} &
0.649 &
0.485 &
0.764 &
0.325 &
0.590 \\

\textbf{MAGMA (ours)} &
0.528 &
\textbf{0.650} &
\textbf{0.517} &
\textbf{0.776} &
\textbf{0.742} &
\textbf{0.700} \\
\bottomrule
\end{tabular}}
\end{table*}

\section{Experiments}
\label{experiments}
We conduct comprehensive experiments to evaluate both the reasoning effectiveness and systems properties of the proposed MAGMA architecture over state-of-the-art baselines.

\subsection{Experimental Setup}
\label{sec:exp_setup}

\noindent\textbf{Datasets.}
We evaluate long-term conversational capability using two widely adopted benchmarks: \textbf{(1) LoCoMo} \citep{maharana2024evaluating}: which contains ultra-long conversations (average length of 9K tokens) designed to assess long-range temporal and causal retrieval. \textbf{(2) LongMemEval} \citep{wu2024longmemeval}: a large-scale stress-test benchmark with an average context length exceeding 100K tokens, used to evaluate scalability and memory retention stability over extended interaction horizons..

\noindent\textbf{Baselines.}
We compare MAGMA against four state-of-the-art memory architectures. For fair comparison, all methods employ the same backbone LLMs.
\begin{itemize}[leftmargin=*, itemsep=0pt, topsep=2pt]
    \item \textbf{Full Context}: Feeds the entire conversation history into the LLM.
    \item \textbf{A-MEM} \citep{xu2025mem}: A biological-inspired, self-evolving memory system that dynamically organizes agent experiences.
    \item \textbf{Nemori} \citep{nan2025nemori}: A graph-based memory utilizing a "predict-calibrate" mechanism for episodic segmentation.
    \item \textbf{MemoryOS}\citep{Kang2025} : A semantic-focused memory operating system employing a hierarchical storage strategy.
\end{itemize}

\noindent\textbf{Metrics.}
Following standard evaluation protocols, we primarily use the LLM-as-a-Judge score~\citep{zheng2023judging} to assess the accuracy of different methods. The detailed evaluation prompt used for the judge model is provided in the appendix. For completeness, we also report token-level F1 and BLEU-1~\citep{papineni-etal-2002-bleu}. 

\subsection{Overall Comparison}
This section introduces the accuracy performance comparison between all methods on the LoCoMo benchmark based on LLM-as-a-judge. As shown in Table~\ref{tab:locomo-main}, MAGMA achieves the highest overall judge score of 0.7, substantially outperforming the other baselines: Full Context (0.481), A-MEM (0.58), MemoryOS (0.553) and Nemori (0.59) by relative margins of 18.6\% to 45.5\%. This result demonstrates that explicitly modeling multi-relational structure enables more accurate long-horizon reasoning than flat or purely semantic memory architectures.

A closer analysis reveals that MAGMA’s advantage is particularly pronounced in reasoning-intensive settings. In the Temporal category, MAGMA slightly but consistently outperforms others (Judge: 0.650 for MAGMA vs.\ 0.422 - 0.649 for others), validating the effectiveness of our Temporal Inference Engine in resolving relative temporal expressions into grounded chronological representations. The performance gap further widens under adversarial conditions, where MAGMA attains a judge score of 0.742. This robustness stems from the Adaptive Traversal Policy, which prioritizes causal and entity-consistent paths and avoids semantically similar yet structurally irrelevant distractors that often mislead vector-based retrieval systems. Additional results and analyzes, including case studies and evaluations under alternative metrics, are provided in the appendix.

\begin{table*}[t]
\centering
\small
\caption{Performance comparison on LongMemEval dataset across different question types. We compare our MAGMA method against the Full-context baseline and the Nemori system.}
\label{tab:longmemeval}
\resizebox{0.8\textwidth}{!}{ 
\begin{tabular}{clccc}
\toprule
& \textbf{Question Type} & \textbf{Full-context} & \textbf{Nemori} & \textbf{MAGMA} \\
& & (101K tokens) & (3.7--4.8K tokens) & (0.7--4.2K tokens) \\
\midrule
\multirow{7}{*}{\rotatebox[origin=c]{90}{\textbf{gpt-4o-mini}}}
& single-session-preference & 6.7\%  & 62.7\% & \textbf{73.3\%} \\
& single-session-assistant  & \textbf{89.3\%} & 73.2\% & 83.9\% \\
& temporal-reasoning        & 42.1\% & 43.0\% & \textbf{45.1\%} \\
& multi-session             & 38.3\% & \textbf{51.4\%} & 50.4\% \\
& knowledge-update          & 78.2\% & 52.6\% & \textbf{66.7\%} \\
& single-session-user       & \textbf{78.6\%} & 77.7\% & 72.9\% \\
& \textit{Average}          & 55.0\% & 56.2\% & \textbf{61.2\%} \\

\bottomrule
\end{tabular}
} 
\end{table*}

\subsection{Generalization Study}
\label{sec:rq2}
To evaluate generalization under extreme context lengths, we compare MAGMA against prior methods on the LongMemEval benchmark. LongMemEval poses a substantial scalability challenge, with an average context length exceeding 100k tokens, and therefore serves as a rigorous stress test for long-term memory retention and retrieval under strict computational constraints. 

As summarized in Table~\ref{tab:longmemeval}, MAGMA achieves the highest average accuracy (61.2\%), outperforming both the Full-context baseline (55.0\%) and the Nemori system (56.2\%). These results indicate that MAGMA generalizes effectively to ultra-long interaction histories while maintaining strong retrieval precision.

At the same time, the results highlight a favorable efficiency–granularity trade-off. Although the Full-context baseline performs strongly on \textit{single-session-assistant} tasks (89.3\%), this performance comes at a prohibitive computational cost, requiring over 100k tokens per query. MAGMA achieves competitive accuracy (83.9\%) while using only 0.7k--4.2k tokens per query, representing a reduction of more than 95\%. This demonstrates that MAGMA effectively compresses long interaction histories into compact, reasoning-dense subgraphs, preserving essential information while substantially reducing inference-time overhead.

\subsection{System Efficiency Analysis}
\label{sec:rq3}

To evaluate the system efficiency of MAGMA, two metrics are focused: (1) memory build time (the time required to construct the memory graph) and (2) token cost (the average tokens processed per query).

Table~\ref{tab:system_perf} reports the comparative results. While A-MEM achieves the lowest token consumption (2.62k) due to its aggressive summarization, it sacrifices reasoning depth (see Table~\ref{tab:locomo-main}). In contrast, MAGMA achieves the lowest query latency (1.47s) about 40\% faster than the next best retrieval baseline (A-MEM) while maintaining a competitive token cost (3.37k). This efficiency stems from our Adaptive Traversal Policy, which prunes irrelevant subgraphs early, and the dual-stream architecture that offloads complex indexing to the background.

\begin{table}[t]
\centering
\small
\setlength{\tabcolsep}{3pt}       
\caption{System efficiency comparison with total memory build time (in hours), average token consumption per query (in k tokens), and average query latency (in seconds).}
\label{tab:system_perf}
\begin{tabular}{l c c c}
\toprule
{Method} & {Build Time (h)} & {Tokens/Query (k)} & {Latency (s)} \\
\midrule
Full Context    & N/A & 8.53 & 1.74 \\
A-MEM           & 1.01 & \textbf{2.62} & 2.26 \\
MemoryOS        & 0.91 & 4.76          & 32.68 \\
Nemori          & \textbf{0.29} & 3.46 & 2.59 \\
\textbf{MAGMA}  & 0.39 & 3.37          & \textbf{1.47} \\ 
\bottomrule
\end{tabular}
\end{table}


\begin{table}[!t]
\centering
\small
\caption{Breakdown analysis on the performance impact of different schemes in MAGMA.}
\label{tab:ablation}
\begin{tabular}{l|ccc}
\toprule
MAGMA schemes & {Judge} & {F1} & {BLEU-1} \\
\midrule
w/o Adaptive Policy & 0.637 & 0.413 & 0.357 \\
w/o Causal Links & 0.644 & 0.439 & 0.354 \\
w/o Temporal Backbone & 0.647 & 0.438 & 0.349 \\
w/o Entity Links & 0.666 & 0.451 & 0.363 \\
\midrule
 \textbf{MAGMA (Full)} & \textbf{0.700} & \textbf{0.467} & \textbf{0.378} \\
\bottomrule
\end{tabular}
\end{table}

\begin{table*}[t]
\centering
\small
\caption{Single-graph ablation study on LoCoMo.}
\label{tab:single_graph_ablation}
\begin{tabular}{l|ccccc|c}
\toprule
Graph Configuration & Multi-Hop & Temporal & Open-Domain & Single-Hop & Adversarial & Overall \\
\midrule
Causal Only   & 0.470 & 0.460 & 0.430 & 0.650 & 0.680 & 0.590 \\
Temporal Only & 0.440 & 0.620 & 0.450 & 0.650 & 0.520 & 0.577 \\
Entity Only   & 0.485 & 0.420 & 0.460 & 0.640 & 0.450 & 0.531 \\
\midrule
\textbf{Full MAGMA} & \textbf{0.528} & \textbf{0.650} & \textbf{0.517} & \textbf{0.776} & \textbf{0.742} & \textbf{0.700} \\
\bottomrule
\end{tabular}
\end{table*}

\subsection{Ablation Study}
\label{sec:rq4}
In this subsection, we conduct a systematic ablation study to assess the contribution of individual components in MAGMA. By selectively disabling edge types and traversal mechanisms, we isolate the sources of its reasoning capability. The results in Table~\ref{tab:ablation} reveal three main findings.

First, removing the Adaptive Policy results in the largest performance drop, with the Judge score decreasing from 0.700 to 0.637. This confirms that intent-aware routing is critical: without it, retrieval degenerates into a static graph walk that introduces structurally irrelevant information and degrades reasoning quality. Second, removing either Causal Links or the Temporal Backbone leads to comparable and substantial performance losses (0.644 and 0.647, respectively), indicating that causal structure and temporal ordering provide complementary, non-substitutable axes of reasoning. Finally, removing Entity Links causes a smaller but consistent decline (0.700 to 0.666), highlighting their role in maintaining entity permanence and reducing hallucinations in entity-centric queries.

To further isolate the contribution of each relation type, we additionally conducted a single-graph-only ablation on LoCoMo, shown in Table~\ref{tab:single_graph_ablation}. The results are consistent with the leave-one-out findings in Table~\ref{tab:ablation}. Among single-graph variants, \textit{Causal Only} achieves the highest overall score (0.590), suggesting that causal relations provide strong logical filtering and robustness against distractor noise. \textit{Temporal Only} performs best on temporal questions (0.620), confirming that explicit temporal structure is particularly important for sequential reasoning. In contrast, \textit{Entity Only} obtains the lowest overall score (0.531): while it remains helpful for concept bridging in multi-hop reasoning, it lacks both timeline awareness and logical filtering, which explains why removing entity links causes the smallest drop in the full-system ablation.

Overall, these results show that no single relation type is sufficient to recover MAGMA's full reasoning capability, as all single-graph variants remain below 0.60 overall. By explicitly decoupling causal, temporal, and entity relations and combining them with adaptive traversal, MAGMA leverages their complementary strengths to achieve the best overall performance.

\section{Conclusion}
\label{conclusion}
We introduced MAGMA, a multi-graph agentic memory architecture that models semantic, temporal, causal, and entity relations within a unified yet disentangled memory substrate. By formulating retrieval as a policy-guided graph traversal and decoupling memory ingestion from asynchronous structural consolidation, MAGMA enables effective long-horizon reasoning while maintaining low inference-time latency. Empirical results on LoCoMo and LongMemEval demonstrate that MAGMA consistently outperforms state-of-the-art memory systems while achieving substantial efficiency gains under ultra-long contexts.



\section{Limitations}
\label{sec:limitations}

While MAGMA demonstrates strong empirical performance, it has several limitations. First, the quality of the constructed memory graph depends on the reasoning fidelity of the underlying Large Language Models used during asynchronous consolidation. This dependency is a shared limitation of agentic memory systems that rely on LLM-based structural inference, as they are susceptible to extraction errors and hallucinations~\citep{pan2024unifying, xi2025rise, wadhwa2023revisiting}. Although MAGMA employs structured prompts and conservative inference thresholds to reduce spurious links, erroneous or missing relations may still arise and propagate to downstream retrieval. Nevertheless, our experimental results indicate that, even under these constraints, agentic memory systems such as MAGMA substantially outperform traditional baselines, including full-context approaches, in long-horizon reasoning tasks.

Second, multi-graph substrate may introduce additional storage and engineering complexity compared to flat, vector-only memory systems. Maintaining multiple relational views and dual-stream processing incurs a little higher implementation and memory overhead, which may limit applicability in highly resource-constrained environments.

Finally, most existing agentic memory systems, including MAGMA, are primarily evaluated on long-context conversational and agentic benchmarks such as LoCoMo and LongMemEval. While these benchmarks effectively stress temporal and causal reasoning, they do not cover the full range of settings in which agentic memory may be required\citep{hu2025memory, jiang2026anatomy}. Extending MAGMA to other scenarios, such as multimodal agents or environments with heterogeneous observation streams, may require additional adaptation and calibration. Addressing these broader evaluation settings remains an important research direction for future work.

\section{Acknowledgment}
This work was partially supported by NSF 2343863, 2413520, 2417747 and 2440611. Any opinions, conclusions, or recommendations expressed in this material are those of the authors and do not necessarily reflect the views of the NSF.

\bibliography{custom}

\appendix

\section{Related Work}
\label{app:relatedwork}
Following the framing in main text, we organize related work along the same progression: from context window extension to retrieval augmented generation (RAG) and finally to memory augmented generation (MAG), and then discuss structured/graph memories and causal reasoning, which are central to long-horizon agentic interactions.\\
\textbf{Context-window Extension.} A direct line of work extends the effective context length of Transformers by modifying attention or positional extrapolation. Longformer~\cite{beltagy2020longformer} introduces sparse attention patterns to scale to long documents, reducing quadratic cost while retaining locality and selected global connectivity. ALiBi~\cite{press2021train} (Attention with Linear Biases) enables length extrapolation by injecting distance-aware linear biases into attention scores, improving robustness when testing on longer sequences than those seen in training. Recent efforts also add explicit memory modules or hybrid mechanisms to push beyond pure attention-window scaling. For example, LM2~\cite{kang2025lm2} proposes a decoder-only architecture augmented with an auxiliary memory to mitigate long-context limitations. MemoRAG~\cite{qian2025memorag} similarly emphasizes global-memory-enhanced retrieval to boost long-context processing when raw context is insufficient or inefficient. While these approaches improve long-range coverage, they do not, by themselves, address the continual, evolving, and write-back nature of agent memory required for multi-session interactions.\\
\textbf{Retrieval Augmented Generation.} RAG~\cite{lewis2020retrieval} augments an LLM with external retrieval over a fixed corpus, classically retrieving supporting passages and conditioning generation on them. Subsequent work explores better integration with long-context models and more scalable retrieval pipelines. LongRAG~\cite{jiang2024longrag} studies how to exploit long-context LLMs together with retrieval, improving the ability to incorporate larger retrieved evidence sets. Other systems focus on structuring the retrieved memory space or optimizing the RAG serving stack: M-RAG~\cite{wang2024m} uses multiple partitions to encourage fine-grained retrieval focus, while RAGO~\cite{jiang2025rago} provides a systematic framework for performance optimization in RAG serving. Furthermore, to improve the reliability of retrieved context, approaches like Self-Correcting RAG~\cite{xu2026selfcorrectingrag} enhance generation faithfulness via selective context filtering and inference-guided search. However, standard RAG typically assumes a static knowledge base. In contrast, agentic settings require memory that is continuously updated (the feedback loop described in the main text). This motivates the shift to MAG systems, where memory is dynamic and evolves with interaction histories.\\
\textbf{Memory Augmented Generation and Agent Memory Systems.} MAG systems maintain and update an external memory over time, enabling agents to accumulate knowledge, preserve identity, and remain coherent across sessions. Early and representative directions include memory construction and write-back strategies for long-term agent behavior, such as MemoryBank~\cite{zhong2024memorybank} and generative agents style architectures that emphasize persistent profiles and evolving state grounded in past interactions~\cite{nan2025nemori,maharana2024evaluating,zeng2025bridging}. A growing body of work adopts systems metaphors and designs: MemGPT~\cite{packer2023memgpt} frames LLM agents with an operating-system-like memory hierarchy, emphasizing paging and controlled context management. More recent memory OS systems propose explicit storage hierarchies, efficient memory modules, and controllers (e.g., MemoryOS~\cite{kang2025memory}, MemOS~\cite{li2025memos}, and Hippocampus~\cite{li2026hippocampus}) to manage persistence, scalability, and retrieval policies at scale. In addition, practical agent-memory stacks (e.g., Zep~\cite{rasmussen2025zep}) offer temporal knowledge-graph-based memory services aimed at real-world deployment constraints.\\
\textbf{Structured memory: chains-of-thought and graph-based representations.} Beyond flat text buffers or vector stores, several methods explicitly structure memory to support reasoning. Think-in-Memory (TiM) stores evolving chains-of-thought to improve consistency across long-horizon reasoning, while A-MEM~\cite{xu2025mem} is inspired by Zettelkasten-style linking of notes/experiences. These methods highlight the value of representing intermediate reasoning traces or explicit links, but many retrieval pipelines still predominantly rely on semantic similarity as the primary access mechanism. Graph-based approaches have recently gained traction as a way to capture cross-document and cross-episode dependencies. GraphRAG~\cite{edge2024local} builds entity-centric graphs and community summaries to answer more global questions over large corpora. Zep proposes a temporally-aware knowledge-graph engine (Graphiti) that synthesizes conversational and structured business data while preserving historical relations. The main text notes these graph-based lines explicitly and motivates a key gap: many systems organize memory around associative proximity (semantic relatedness) rather than mechanistic dependency.\\
\textbf{Causal reasoning and long-horizon evaluation.} Causal reasoning has been highlighted as both important and challenging for LLMs. The work~\cite{kiciman2023causal} study LLMs’ ability to generate causal arguments across multiple causal tasks and emphasize robustness/failure modes, reinforcing that what happened retrieval is not sufficient for why reasoning in many settings. Benchmarking efforts such as LoCoMo~\cite{maharana2024evaluating} stress long-range temporal and causal dynamics in multi-session conversations and provide evaluation tasks that expose long-horizon memory deficits. The paper’s experimental setup also uses LongMemEval~\cite{wu2024longmemeval} as an ultra-long context stress test, and evaluates via LLM-as-a-Judge protocols standard in modern instruction-following evaluation. Overall, prior work demonstrates steady progress in (i) scaling context length, (ii) improving retrieval pipelines, and (iii) building structured, evolving memories for agents. The main text positions MAGMA within this trajectory by explicitly targeting multi-relational structure (semantic/temporal/causal/entity) and intent-aware retrieval control.

\section{System Implementation Details}
\label{app:impl}

\subsection{Hyperparameter Configuration}
Table~\ref{tab:hyperparams} presents the comprehensive configuration used in our experiments. These parameters were empirically optimized on the LoCoMo benchmark. Notably, MAGMA employs an \textit{Adaptive Scoring} mechanism where weights ($\lambda$) shift dynamically based on the detected query intent.

\begin{table}[h]
\centering
\footnotesize
\caption{Hyperparameter settings for MAGMA. "Traversal Weights" correspond to the intent-specific vector $\mathbf{w}_{T_q}$, while $\lambda_{1}$ and $\lambda_{2}$ control the global balance between structural alignment and semantic affinity (Eq.~\ref{eq:beam_score}).}
\label{tab:hyperparams}
\resizebox{\linewidth}{!}{
\begin{tabular}{l l l}
\toprule
\textbf{Module} & \textbf{Parameter} & \textbf{Value/Range} \\
\midrule
\multirow{3}{*}{Embedding} 
& Model (Default) & \texttt{all-MiniLM-L6-v2} \\
& Model (Optional) & \texttt{text-embedding-3-small} \\
& Dimension & 384 / 1536 \\
\midrule
\multirow{2}{*}{Inference} 
& LLM Backbone & \texttt{gpt-4o-mini} \\
& Temperature & 0.0 \\
\midrule
\multirow{4}{*}{Retrieval (Phase 1)} 
& RRF Constant ($k$) & 60 \\
& Vector Top-K & 20 \\
& $w_{keyword}$ (Fusion) & 2.0 -- 5.0 \\ 
& Sim. Threshold & 0.10--0.30 \\
\midrule
\multirow{3}{*}{Traversal (Phase 2)}
& Max Depth & 5 hops \\
& Max Nodes & 200 \\
& Drop Threshold & 0.15 \\
\midrule
\multirow{6}{*}{\shortstack[l]{Adaptive Weights}} 
& $\lambda_{1}$ (Structure Coef.) & 1.0 (Base) \\
& $\lambda_{2}$ (Semantic Coef.) & 0.3 -- 0.7 \\
\cmidrule(l){2-3} 
& $w_{entity}$ (in $\mathbf{w}_{T_q}$) & 2.5 -- 6.0 \\
& $w_{temporal}$ (in $\mathbf{w}_{T_q}$) & 0.5 -- 4.0 \\
& $w_{causal}$ (in $\mathbf{w}_{T_q}$) & 3.0 -- 5.0 \\
& $w_{phrase}$ (in $\mathbf{w}_{T_q}$) & 2.5 -- 5.0 \\
\bottomrule
\end{tabular}
}
\end{table}

\section{Prompt Library}
\label{app:prompts}

MAGMA employs a sophisticated prompt strategy with three distinct types, each optimized for specific cognitive tasks within the memory pipeline.

\subsection{Event Extraction Prompt (JSON-Structured)}
To ensure robustness against hallucination and parsing errors, this module employs a strict JSON schema enforcement strategy. The prompt explicitly defines the extraction targets to ensure downstream graph integrity, capturing not just entities but also semantic relationships and temporal markers.

\begin{tcolorbox}[colback=gray!10, colframe=black, title=\textbf{System Prompt: Event Extractor}, breakable]
\small
\raggedright
\textbf{System Role:} You are an automated Graph Memory Parser. Your task is to extract structured metadata from raw conversational logs to build a knowledge graph.

\textbf{Input Data:}
\begin{itemize}[nosep, leftmargin=*]
    \item Speaker: \{speaker\}
    \item Text: \{text\}
    \item Context: \{prev\_summary\}
\end{itemize}

\textbf{Instructions:}
Analyze the input and return \textbf{ONLY} a valid JSON object matching the specific schema below. Do not include markdown formatting.

\textbf{Target Schema:}
\begin{itemize}[nosep, leftmargin=*]
    \item \texttt{"entities"}: List of proper nouns (People, Locations, Organizations).
    \item \texttt{"topic"}: String (1--3 words representing the main theme).
    \item \texttt{"relationships"}: List of strings describing interactions (e.g., "X researches Y").
    \item \texttt{"semantic\_facts"}: List of atomic facts preserving key information.
    \item \texttt{"dates\_mentioned"}: List of temporal strings (e.g., "next Friday", "2024-01-01").
    \item \texttt{"summary"}: One-sentence summary preserving speaker attribution.
\end{itemize}
\end{tcolorbox}

\subsection{Query-Adaptive QA Prompt}
The generation prompt begins with a strict persona definition and appends specific reasoning instructions dynamically based on the Router's classification (e.g., Multi-hop, Temporal, Open-domain).

\begin{tcolorbox}[colback=gray!10, colframe=black, title=\textbf{System Prompt: Adaptive QA}, breakable]
\small
\raggedright
\textbf{System Role:} You are a precision QA assistant operating on retrieved memory contexts. Your goal is to answer the user's question accurately using \textit{only} the provided information.

\textbf{Context:}
\{context\}

\textbf{Current Query:}
\begin{itemize}[nosep, leftmargin=*]
    \item Question: \{question\}
    \item Constraints: \{category\_specific\_constraints\}
\end{itemize}

\textbf{Instructions:}
\begin{enumerate}[nosep, leftmargin=*]
    \item Use ONLY information explicitly stated in the context.
    \item If the answer is not present, respond exactly with "Information not found".
    \item Be concise (typically 1--10 words) unless detailed reasoning is required.
    \item \textbf{\{dynamic\_instruction\}} \textcolor{gray!80}{\textit{// Automatically generated by our engine's query classifier/router (no oracle labels)}}

\end{enumerate}

\textbf{Answer:} 

\tcblower 

\textbf{*Dynamic Instruction Injection Candidates:}
\begin{itemize}[nosep, leftmargin=*]
    \item \textbf{[Multi-hop]:} "Connect related facts across different nodes. For comparison queries (e.g., 'both/all'), identify commonalities between entities rather than listing individual details."
    
    \item \textbf{[Temporal]:} "Resolve relative dates (e.g., 'yesterday') using the event timestamps. Output dates strictly in 'D Month YYYY' format. Calculate durations if asked."
    
    \item \textbf{[Open-Domain/Inference]:} "Make reasonable inferences based on the user's personality traits, interests, and past behaviors. Support hypothetical ('would/could') reasoning with evidence."
    
    \item \textbf{[Single-hop/Factual]:} "Extract the specific entity, name, or method requested. Do not add explanations. Return the exact fact matching the query intent."
\end{itemize}
\end{tcolorbox}

\subsection{Evaluation Prompt (LLM-as-a-Judge)}
To ensure rigorous evaluation beyond simple n-gram overlapping, we employ a semantic scoring mechanism. The Judge LLM evaluates the alignment between the generated response and the ground truth using the following schema.

\begin{tcolorbox}[colback=gray!10, colframe=black, title=\textbf{System Prompt: Semantic Grader}, breakable]
\small
\raggedright
You are an expert evaluator assessing the semantic fidelity of a memory retrieval system. Score the \texttt{Candidate Answer} against the \texttt{Gold Reference} on a continuous scale [0.0, 1.0].

\textbf{Scoring Rubric:}
\begin{itemize}[nosep, leftmargin=*]
    \item \textbf{1.0 (Exact Alignment):} Captures all key entities, temporal markers, and causal relationships. Semantically equivalent.
    \item \textbf{0.8 (Substantially Correct):} Main point is accurate but lacks minor nuances or secondary details.
    \item \textbf{0.6 (Partial Match):} Contains valid information but misses key constraints (e.g., wrong date but correct event).
    \item \textbf{0.4 (Tangential):} Touches on the topic but misses the core information requirement.
    \item \textbf{0.2 (Incoherent):} Factually incorrect with only minimal topical overlap.
    \item \textbf{0.0 (Contradiction/Hallucination):} Completely unrelated or contradicts the ground truth.
\end{itemize}

\textbf{Evaluation Constraints:}
\begin{enumerate}[nosep, leftmargin=*]
    \item \textbf{Temporal Flexibility:} Accept relative time references (e.g., "next Tuesday") if they resolve to the same period as the Gold Reference.
    \item \textbf{Semantic Equivalence:} Prioritize informational content over lexical matching.
    \item \textbf{Adversarial Handling:} If the Gold Reference states "Unanswerable", the Candidate MUST explicitly state lack of information. Any hallucinated fact results in 0.0.
\end{enumerate}

\textbf{Input:} Question: \{question\} | Gold: \{gold\} | Candidate: \{generated\}\\
\textbf{Output:} JSON \texttt{\{"score": float, "reasoning": "concise explanation"\}}
\end{tcolorbox}

\section{Baseline Configurations}
\label{app:baselines}

To ensure a fair and rigorous comparison, we standardized the experimental environment across all systems. Specifically, we adhered to the following protocols:

\begin{itemize}
    \item \textbf{Full Context Baseline:} We implemented a "Full Context" baseline where the entire available conversation history is fed directly into the LLM's context window (up to the 128k token limit of \texttt{gpt-4o-mini}). This serves as a "brute-force" reference to evaluate the model's native long-context capabilities without external retrieval mechanisms.
    \item \textbf{Retrieval-Based Baselines:} For all baseline systems (e.g., AMem, Nemori, MemoryOS), we applied their official default hyperparameters and storage settings to reflect their standard out-of-the-box performance.
    \item \textbf{Unified Backbone Model:} To eliminate performance variance caused by different foundation models, all systems utilized OpenAI's \texttt{gpt-4o-mini} for both retrieval reasoning and response generation.
    \item \textbf{Unified Evaluation:} All system outputs were evaluated using the identical \textit{LLM-as-a-Judge} framework (also powered by \texttt{gpt-4o-mini} with temperature=0.0), as detailed in Appendix~\ref{app:prompts}.
\end{itemize}

\paragraph{Dataset Statistics.}
We conducted a comprehensive evaluation on the full LoCoMo benchmark, testing across all five cognitive categories to assess varying levels of retrieval complexity. The detailed distribution of query types is presented in Table~\ref{tab:locomo_stats}.

\begin{table}[h]
\centering
\small
\caption{Distribution of query categories in the LoCoMo benchmark used for evaluation.}
\label{tab:locomo_stats}
\begin{tabular}{l r}
\toprule
\textbf{Query Category} & \textbf{Count} \\
\midrule
Single-Hop Retrieval & 841 \\
Adversarial & 446 \\
Temporal Reasoning & 321 \\
Multi-Hop Reasoning & 282 \\
Open Domain & 96 \\
\midrule
\textbf{Total Samples} & \textbf{1,986} \\
\bottomrule
\end{tabular}
\end{table}

\section{Case Study}
\label{sec:case_study}

To demonstrate MAGMA's reasoning capabilities across different cognitive modalities, we analyze three real-world scenarios from the LoCoMo benchmark. Table~\ref{tab:case_study_comparison} provides a side-by-side comparison of MAGMA against key baselines (A-MEM, Nemori, MemoryOS).

\begin{table*}[t]
\centering
\small
\renewcommand{\arraystretch}{1.5}
\caption{Case study for failure analysis comparing MAGMA against baselines across three reasoning types. \textcolor{red}{Red text} indicates hallucinations or partial failures; \textcolor{teal}{\textbf{Teal text}} indicates correct reasoning derived from graph traversal.}
\label{tab:case_study_comparison}

\begin{tabularx}{\textwidth}{p{0.02\textwidth} p{0.18\textwidth} | X | X }
\toprule
& \textbf{Query \& Type} & \textbf{Baseline Failure Mode} & \textbf{MAGMA Graph Reasoning (Success)} \\
\midrule

\rotatebox[origin=c]{90}{\textbf{Fact}} & 
\textbf{Q1: Fact Retrieval} \newline \textit{"What instruments does Melanie play?"} & 
\textbf{A-MEM:} \textcolor{red}{"Memories do not explicitly state..."} \newline
\textbf{MemoryOS:} "Clarinet" \newline
\textit{\textbf{Failure:}} Baselines relying on top-k vector search missed the distant memory of the "violin" (D2:5) because it appeared in a context about "me-time" rather than explicitly about music. & 
\textcolor{teal}{\textbf{"Clarinet and Violin."}} \newline
\textit{\textbf{Mechanism:}} MAGMA utilized the entity-centric subgraph around "Melanie". By traversing dynamic semantic edges (e.g., "playing", "enjoy") to connected event nodes, it aggregated all mentions of musical activities regardless of the specific relation label or distance. \\
\hline

\rotatebox[origin=c]{90}{\textbf{Logic}} & 
\textbf{Q2: Logical Inference} \newline \textit{"How many children does Melanie have?"} & 
\textbf{Nemori:} \textcolor{red}{"At least two..."} \newline
\textbf{MemoryOS:} \textcolor{red}{"Two"} \newline
\textit{\textbf{Failure:}} Baselines performed surface-level extraction from a photo description showing "two children" (D18:5), failing to account for the "son" mentioned in a separate accident event. & 
\textcolor{teal}{\textbf{"At least three."}} \newline
\textit{\textbf{Mechanism:}} MAGMA executed multi-hop inference focused on \textbf{Entity Resolution}:
1. Node A (Photo): Identified "two kids" entity.
2. Node B (Accident): Linked "son" (D18:1) via a dynamic relationship edge.
3. Node C (Dialogue): Confirmed "brother" (D18:7) is distinct from the two in the photo.
$\rightarrow$ Logic: $2 \text{ (Photo)} + 1 \text{ (Son/Brother)} = 3$. \\
\hline

\rotatebox[origin=c]{90}{\textbf{Time}} & 
\textbf{Q3: Temporal Res.} \newline \textit{"When did she hike after the roadtrip?"} & 
\textbf{A-MEM:} \textcolor{red}{"20 October 2023"} \newline
\textbf{MemoryOS:} \textcolor{red}{"29 December 2025"} \newline
\textit{\textbf{Failure:}} A-MEM simply copied the session timestamp. MemoryOS hallucinated a future date. Both failed to resolve the relative time expression. & 
\textcolor{teal}{\textbf{"19 October 2023"}} \newline
\textit{\textbf{Mechanism:}} MAGMA's Temporal Parser identified the relative marker "yesterday" in D18:17.
Calculation: $T_{\text{session}} (Oct 20) - 1 \text{ day} = Oct 19$.
This exact date was anchored to the Event Node, allowing precise retrieval. \\
\bottomrule
\end{tabularx}
\end{table*}

\subsection{Illustrative Walkthrough: From Memory Construction to Retrieval}
To make the case study more concrete, we briefly illustrate how MAGMA processes the Melanie example end to end. Consider three representative facts from the conversation history: (1) Melanie mentions playing the \textit{violin} in an earlier session, (2) she later says that she also plays the \textit{clarinet}, and (3) in a later family-trip session, she refers to her \textit{son}, \textit{two children} in a photo, and a hike done \textit{yesterday}. 

During memory construction, MAGMA first segments these utterances into event nodes and places them along a temporal backbone. The system then incrementally enriches this memory with additional structure, such as semantic connections between related musical events, entity-centric links for recurring references to Melanie and her family members, and normalized temporal attributes for relative expressions such as \textit{yesterday}. As a result, the memory is not stored as a flat list of text snippets, but as a small multi-view graph in which the same history can be accessed through different relational paths.

At query time, MAGMA first identifies the dominant retrieval intent and then selects anchor events before traversing the most relevant graph views. For example, for \textit{``What instruments does Melanie play?''}, retrieval focuses on the entity and semantic views, allowing MAGMA to aggregate both the earlier \textit{violin} mention and the later \textit{clarinet} mention. For \textit{``How many children does Melanie have?''}, retrieval centers on the local entity neighborhood, combining the references to a \textit{son}, \textit{two children}, and \textit{brother} into a single evidence set. For \textit{``When did she hike?''}, MAGMA relies primarily on the temporal view, using the normalized representation of \textit{yesterday} to recover the grounded date.

This example highlights the key intuition behind MAGMA: memory construction organizes conversational history into complementary relational views, and query-time retrieval activates different parts of this structure depending on the reasoning need.

\subsection{Detailed Analysis}

\paragraph{Case 1: Overcoming Information Loss (Recall).}
For the query regarding instruments, A-MEM failed completely due to its summarization process abstracting away specific details ("violin") from early sessions. Other RAG baselines only retrieved the "clarinet" due to surface-level semantic matching. MAGMA, however, maintains an entity-centric graph structure. Instead of relying on rigid schemas, MAGMA queries the local neighborhood of the \texttt{[Entity: Melanie]} node. This allows it to capture diverse natural language predicates (e.g., "playing my violin", "started clarinet") and aggregate disjoint facts into a comprehensive answer, demonstrating robustness against information loss.

\paragraph{Case 2: Multi-Hop Reasoning vs. Surface Extraction.}
The query "How many children?" exposes a critical weakness in standard RAG: the inability to perform arithmetic across contexts. Baselines simply extracted the explicit mention of "two children" from a photo caption. In contrast, MAGMA treated this as a graph traversal problem focused on entity resolution. It queried the neighborhood of \texttt{[Entity: Melanie]} for connected nodes of type \texttt{Person}. By analyzing the semantic edges, specifically distinguishing the "two kids" entity in the canyon photo from the "son" entity involved in the car accident, MAGMA synthesized these distinct nodes. It correctly deduced that the "son" (referenced later as "brother") was an additional individual, summing up to a count of "at least three," a logical leap impossible for systems relying solely on vector similarity.

\paragraph{Case 3: Temporal Grounding.}
When asked "When did she hike?", baselines either hallucinated or defaulted to the conversation timestamp (Oct 20). This ignores the semantic meaning of the user's statement: "we just did it \textit{yesterday}." MAGMA's structured ingestion pipeline normalizes relative dates during graph construction. The event was stored with the resolved attribute \texttt{date="2023-10-19"}, making the retrieval trivial and exact, completely bypassing the ambiguity that confused the LLM-based baselines.

\begin{table*}[t]
\centering
\small
\caption{LoCoMo evaluation with F1 and BLEU-1 metrics}
\label{tab:locomo-f1}

\resizebox{\textwidth}{!}{
\begin{tabular}{lcccccccc|cccc}
\toprule
& \multicolumn{2}{c}{Multi-Hop} &
  \multicolumn{2}{c}{Temporal} &
  \multicolumn{2}{c}{Open-Domain} &
  \multicolumn{2}{c}{Single-Hop} &
  \multicolumn{2}{|c}{Overall} \\
\cmidrule(lr){2-3}\cmidrule(lr){4-5}\cmidrule(lr){6-7}
\cmidrule(lr){8-9}\cmidrule(lr){10-11}
Method & F1 & BLEU-1 &
         F1 & BLEU-1 &
         F1 & BLEU-1 &
         F1 & BLEU-1 &
         F1 & BLEU-1 \\
\midrule
\midrule
Full Context &
0.182 & 0.128 &
0.079 & 0.055 &
0.042 & 0.030 &
0.229 & 0.156 &
0.140 & 0.096 \\

A-MEM &
0.128 & 0.088 &
0.128 & 0.079 &
0.076 & 0.051 &
0.174 & 0.110 &
0.116 & 0.074 \\

MemoryOS &
0.365 & 0.276 &
0.434 & 0.369 &
0.246 & 0.191 &
0.493 & 0.437 &
0.413 & 0.355 \\

Nemori &
0.363 & 0.249 &
0.569 & 0.479 &
0.247 & 0.189 &
0.548 & 0.439 &
0.502 & 0.403 \\

MAGMA (ours) &
0.264 & 0.172 &
0.509 & 0.370 &
0.180 & 0.136 &
0.551 & 0.477 &
0.467 & 0.378 \\
\bottomrule
\end{tabular}}
\end{table*}

\section{Metric Validation Analysis}
\label{app:metric_validation}

To validate our choice of using an LLM-based Judge over traditional lexical metrics, we conducted a granular failure analysis on seven representative test cases. Table~\ref{tab:metric_cases} details the quantitative breakdown.

\subsection{Rationale for Semantic Scoring}
Our empirical results reveal two critical failure modes where standard metrics (F1, BLEU-1) directly contradict human judgment:

\begin{enumerate}
    \item \textbf{False Rewards} (The ``Hallucination'' Problem): Lexical metrics heavily reward incorrect answers that share surface-level tokens. 
    \begin{itemize}
        \item In Case 3, a direct negation (``compatible'' vs. ``\textit{not} compatible'') yields a remarkably high F1 of 0.857, treating a fatal contradiction as a near-perfect match.
        \item In Case 6, substituting the wrong entity (``John'' vs. ``Sarah'') still achieves \textbf{F1 0.750}, rewarding the hallucinatory output.
    \end{itemize}

    \item \textbf{False Penalties} (The ``Phrasing'' Problem): Valid answers with different formatting or synonyms are unfairly penalized.
    \begin{itemize}
        \item In Case 4 (Time Notation) and Case 5 (Synonyms), F1 and BLEU scores drop to 0.000 despite the answers being semantically identical.
    \end{itemize}
\end{enumerate}

As shown in Table~\ref{tab:metric_cases}, the LLM-Judge correctly assigns a score of 0.0 to factual errors and 1.0 to semantic matches, aligning perfectly with reasoning requirements.

\begin{table*}[h]
\centering
\caption{Quantitative Failure Analysis of Lexical Metrics. We present seven controlled cases with their calculated F1 and BLEU-1 scores. The data demonstrates that lexical metrics frequently assign high scores to fatal errors (False Rewards) and zero scores to correct variations (False Penalties), whereas the LLM-Judge correctly assesses semantic validity.}
\label{tab:metric_cases}

\begin{tcolorbox}[colback=gray!5, colframe=gray!60, arc=2mm, boxrule=0.8pt]
\small
\renewcommand{\arraystretch}{1.6} 
\begin{tabularx}{\textwidth}{p{0.25\textwidth} | X | p{0.12\textwidth} p{0.12\textwidth}}
\toprule
\rowcolor{gray!20} 
\textbf{Failure Mode} & \textbf{Case Detail (Gold / Predicted)} & \textbf{Lexical Metrics} \newline \textit{(F1 / BLEU-1)} & \textbf{LLM Judge} \newline \textit{(Semantic)} \\ 
\midrule

\textbf{Case 1: False Reward} \newline \textit{(Wrong Fact, High Overlap)} & 
\textbf{Gold:} ``three items'' \newline
\textbf{Pred:} ``five items'' \newline
\textit{\textbf{Analysis:}} Factually wrong count, but rewarded for sharing the noun ``items''. & 
\textcolor{red}{\textbf{High}} \newline 0.500 / 0.500 & 
\textcolor{teal}{\textbf{0.0}} \newline (Reject) \\ 
\hline

\textbf{Case 2: False Penalty} \newline \textit{(Verbose Phrasing)} & 
\textbf{Gold:} ``18 days'' \newline
\textbf{Pred:} ``The total duration was 18 days'' \newline
\textit{\textbf{Analysis:}} Correct answer penalized for low precision due to extra words. & 
\textcolor{red}{\textbf{Low}} \newline 0.500 / 0.333 & 
\textcolor{teal}{\textbf{1.0}} \newline (Accept) \\ 
\hline

\textbf{Case 3: False Reward} \newline \textit{(Negation/Contradiction)} & 
\textbf{Gold:} ``compatible with Mac'' \newline
\textbf{Pred:} ``\textbf{not} compatible with Mac'' \newline
\textit{\textbf{Analysis:}} Fatal contradiction receives near-perfect scores due to high token overlap. & 
\textcolor{red}{\textbf{Very High}} \newline 0.857 / 0.750 & 
\textcolor{teal}{\textbf{0.0}} \newline (Reject) \\ 
\hline

\textbf{Case 4: False Penalty} \newline \textit{(Time Notation)} & 
\textbf{Gold:} ``14:00'' \newline
\textbf{Pred:} ``2 PM'' \newline
\textit{\textbf{Analysis:}} Different formats result in zero overlap despite identical meaning. & 
\textcolor{red}{\textbf{Zero}} \newline 0.000 / 0.000 & 
\textcolor{teal}{\textbf{1.0}} \newline (Accept) \\ 
\hline

\textbf{Case 5: False Penalty} \newline \textit{(Synonyms)} & 
\textbf{Gold:} ``cheap'' \newline
\textbf{Pred:} ``inexpensive'' \newline
\textit{\textbf{Analysis:}} Standard metrics cannot handle synonym matching without external resources. & 
\textcolor{red}{\textbf{Zero}} \newline 0.000 / 0.000 & 
\textcolor{teal}{\textbf{1.0}} \newline (Accept) \\ 
\hline

\textbf{Case 6: False Reward} \newline \textit{(Entity Hallucination)} & 
\textbf{Gold:} ``John completed the project'' \newline
\textbf{Pred:} ``Sarah completed the project'' \newline
\textit{\textbf{Analysis:}} Wrong entity (Sarah vs John), yet high metrics due to shared sentence structure. & 
\textcolor{red}{\textbf{High}} \newline 0.750 / 0.750 & 
\textcolor{teal}{\textbf{0.0}} \newline (Reject) \\ 
\hline

\textbf{Case 7: False Penalty} \newline \textit{(Format Noise)} & 
\textbf{Gold:} ``5'' \newline
\textbf{Pred:} ``5 (extracted from JSON...)'' \newline
\textit{\textbf{Analysis:}} Correct value embedded in noise results in poor precision metrics. & 
\textcolor{red}{\textbf{Low}} \newline 0.286 / 0.167 & 
\textcolor{teal}{\textbf{1.0}} \newline (Accept) \\ 
\bottomrule

\end{tabularx}
\end{tcolorbox}
\end{table*}

\end{document}